\documentclass[letterpaper, 10 pt, conference]{ieeeconf}
\IEEEoverridecommandlockouts
\usepackage{cite}
\usepackage{amsmath,amssymb,amsfonts}
\usepackage{algorithmic}
\usepackage{graphicx}
\usepackage{textcomp}
\usepackage{xcolor}
\def\BibTeX{{\rm B\kern-.05em{\sc i\kern-.025em b}\kern-.08em
    T\kern-.1667em\lower.7ex\hbox{E}\kern-.125emX}}
\begin{document}

\title{Leveraging GCN-based Action Recognition for Teleoperation in Daily Activity Assistance
}

\author{
Thomas M. Kwok$^{1}$, Jiaan Li$^1$, Yue Hu$^1$
\thanks{$^{1}$ Department of Mechanical and Mechatronics Engineering, University of Waterloo, Canada {\tt\footnotesize thomasm.kwok@uwaterloo.ca, j232li@uwaterloo.ca, yue.hu@uwaterloo.ca}}
\thanks{This work was supported
in part by the National Research Council (NRC) under Grant NRC-AiP-302-1; and the support of Waterloo RoboHub for the equipment used. We acknowledge the support of the Natural Sciences and Engineering Research Council of Canada (NSERC), funding reference number RGPIN-2022-03857.}
}

\maketitle

\begin{abstract}
Caregiving of older adults is an urgent global challenge, with many older adults preferring to age in place rather than enter residential care. However, providing adequate home-based assistance remains difficult, particularly in geographically vast regions. Teleoperated robots offer a promising solution, but conventional motion-mapping teleoperation imposes unnatural movement constraints on operators, leading to muscle fatigue and reduced usability. This paper presents a novel teleoperation framework that leverages action recognition to enable intuitive remote robot control. Using our simplified Spatio-Temporal Graph Convolutional Network (S-ST-GCN), the system recognizes human actions and executes corresponding preset robot trajectories, eliminating the need for direct motion synchronization. A finite-state machine (FSM) is integrated to enhance reliability by filtering out misclassified actions. Our experiments demonstrate that the proposed framework enables effortless operator movement while ensuring accurate robot execution. This proof-of-concept study highlights the potential of teleoperation with action recognition for enabling caregivers to remotely assist older adults during activities of daily living (ADLs). Future work will focus on improving the S-ST-GCN's recognition accuracy and generalization, integrating advanced motion planning techniques to further enhance robotic autonomy in older adult care, and conducting a user study to evaluate the system's telepresence and ease of control.
\end{abstract}


\section{Introduction}

Caregiving of older adults is becoming an increasingly pressing global issue. In countries like Canada, the demand for older adults' caregiving services is rapidly growing. According to Statistics Canada \cite{sc_canada_population}, in 2024, 18.9\% of the population is aged 65 or older—an increase of 3.3\% over the past decade. This proportion is expected to reach 25\% by 2051 \cite{nrc_canada_population}. However, this trend is not unique to Canada. The aging population is a widespread phenomenon. For instance, Monaco's population reached 36\% older adults in 2024, while Japan and Italy have 29\% and 24\% older adult populations, respectively \cite{prb_ageing}. The global aging trend is projected to continue, with the World Health Organization \cite{noauthor_ageing_nodate} forecasting that by 2050, 22\% of the global population (approximately 2.1 billion people) will be over 60 years old.

Most older adults prefer to age in place and maintain their independence at home, rather than transitioning to residential care \cite{nrc_canada_population}. However, due to a decline in muscle strength and mobility \cite{ghozzi_fatigue_2024, torossian_chronic_2021, belkacem_brain_2020}, many older adults require physical assistance with daily activities of living (ADLs), which is one of the main reasons for moving into residential care \cite{age_in_place_cihi}. Interestingly, \cite{age_in_place_cihi} reports that 22\% of older adults who entered residential care could have remained at home with the appropriate support.

Providing support for the older adult population has become a significant challenge and social burden, especially in countries with vast geographical landscapes. In rural areas, home care options are limited, and caregivers often face long commutes. In Canada, individuals living in rural areas are more than 50\% more likely to enter residential care than those in urban areas, even when their conditions would allow them to remain at home \cite{home_care_potential_cihi}. To address this geographical challenge, teleoperated robots, supported by advanced 5G/6G telecommunications \cite{dang_what_2020}, offer a promising solution for remotely caring for the older adults \cite{lyu_teleoperation_2020, shah_toward_2024, meccanici_probabilistic_2023, begum_performance_2013}.

Motion mapping \cite{lyu_teleoperation_2020, peuchpen_real_time_2024, darvish_teleoperation_2023, lin_shared_2020} is commonly used in teleoperation due to its simplicity and intuitive operation. However, during teleoperation, the operator's motion may differ from in-person operation, resulting in unnatural speed and posture. Operators often need to slow down their movements to match the robot’s due to hardware limitations and safety concerns. Research by \cite{lin_physical_2019} has shown that operators experience muscle fatigue, particularly during tasks requiring precise manipulation and steady posture maintenance for activities of daily living (ADLs). As a result, without more effective teleoperation techniques to assist older adults in ADLs, long-term care becomes a significant burden on caregivers, leading to muscle fatigue and increased risks to occupational health.

Another drawback of motion-mapping teleoperation is that the robot's motion is kinematically constrained by the operator’s actions. As a result, the robot’s movements may not be optimized to efficiently complete the task given its configuration, which is different from humans. For example, some studies use shared autonomous interfaces to assist operators with object grasping \cite{lin_shared_2020} or to help operators complete ADL tasks by estimating their motion intent \cite{jun_jeon_shared_2020}. While this approach alleviates the challenges of precise tele-manipulation, the robot can only proceed and finish the task after the operator has indicated their intent via motion-mapping teleoperation. The robot's motion may not be optimized for task completion, particularly after it follows the operator's motion at the early stage.

Today, with advanced robotic autonomy frameworks, such as reinforcement learning combined with diffusion policy learning from demonstrations \cite{wang_diffusion_2022, wang_diffusion_2023, chi_diffusion_2024}, robots can autonomously complete complex tasks without operator involvement, provided they understand human action commands. Some previous work has focused on generating robot motion using high-level action commands. For instance, a voice command like "move to" can direct the robot to perform contextual movements \cite{zhang_llm_driven_2025}. However, a simple command like "move to" may not be sufficient for more intricate ADL tasks, such as food preparation involving cutting and stabbing motions. Another avenue of research explores using Large Language Models for task planning \cite{shin_task_2024}, which can accommodate more complex actions. Yet, this approach may be cumbersome for caregivers who need to describe a sequence of actions in a caregiving setting, and the verbal commands might also confuse older adults. Directly performing the actions would be a more intuitive and straightforward approach for caregivers.

Our goal is to develop a novel teleoperation framework utilizing action recognition, enabling remote robots to understand and autonomously execute human actions. In this proof-of-concept paper, we focus on demonstrating the feasibility of incorporating action recognition into the teleoperation framework as the first step of development. For this purpose, we use preset robot trajectories to carry out the actions. To achieve action recognition, we employ a simplified Spatio-Temporal Graph Convolutional Network (S-ST-GCN), which is a Graph Convolutional Network (GCN) model that incorporates both spatial and temporal structures. Once a human action is recognized, the robot executes the corresponding action using the appropriate preset trajectory. This framework potentially enables caregivers to remotely control the robot in performing ADLs, thereby assisting older adults in maintaining independence at home.

The rest of the paper is organized as follows. Section II
describes the action recognition algorithm using a GNN model and its implementation into a teleoperation framework. Section III presents the experimental validation of the action recognition algorithm and the proposed teleoperation frameworks.
Section IV contains a discussion and potential research. Finally, Section V concludes the contribution of the paper.

\section{Teleoperation with Action Recognition}

\subsection{Action Recognition}

Human action recognition (HAR) has been a prominent research topic for many years. Various approaches have been explored for action recognition or classification, utilizing different types of sensors. For example, action recognition based on human motion with accelerometers \cite{bao_activity_2004}, heart activity with electrocardiograms (ECG) \cite{4650398}, muscle activation with electromyography (EMG) \cite{li_surface_2020}, and environmental factors with light and temperature sensors \cite{6365160}. However, these methods can be impractical for caregivers due to the long preparation and calibration times required before teleoperation. Additionally, the quality of the sensing signals and the accuracy of action recognition may be highly sensitive to sensor placement, further complicating real-time applications.

Vision-based human action recognition (HAR) is one of the most commonly used approaches due to its ease of installation and minimal preparation time. Many existing studies apply pattern recognition techniques such as Support Vector Machines (SVM) \cite{Action_Danafar2007}, k-Nearest Neighbors (k-NN) \cite{1544882}, Random Forests \cite{4587628}, and Convolutional Neural Networks (CNNs) \cite{li_research_2024, kong_human_2022}. However, these methods often struggle in dynamic environments, as they tend to focus on specific video frame features, such as edges. This makes them more suitable for controlled settings with known environments, fixed objects, and operators wearing similar outfits. As a result, implementing these algorithms in real-world caregiving scenarios can be impractical, as the conditions are more varied and unpredictable.

Apart from spatial information, models that capture temporal information from video sequences have shown improved recognition accuracy. Techniques like Recurrent Neural Networks (RNNs) \cite{7298714} and Convolutional Neural Networks (CNNs) combined with Long Short-Term Memory (LSTM) networks \cite{zhen_highly_2023} are commonly used for this purpose. Since actions unfold over time, capturing dependencies between frames enhances human action recognition (HAR) beyond the spatial information available in a single frame.

Building on these considerations, we aim to develop a teleoperation framework with action recognition that leverages RGB cameras along with spatial-temporal human skeleton data.

\subsection{Simplified Spatio-Temporal Graph Convolutional Network (S-ST-GCN)}

Inspired by Spatial-Temporal Graph Convolutional Networks (ST-GCN) \cite{yan_spatial_2018}, we designed a simplified Spatial-Temporal Graph Neural Network (S-ST-GCN) for teleoperation in this proof-of-concept paper. ST-GCN \cite{yan_spatial_2018} utilizes both spatial and temporal information of human skeletons for action recognition, making it a promising approach for teleoperation in caregiving settings. This method is vision-based, eliminating the need for additional remote controllers or sensors, thereby simplifying the setup. Relying on human skeleton data instead of specific video frame features like edges, it offers robustness in action recognition, accommodating different operators and varying environments.

In \cite{yan_spatial_2018}, Yan introduced a graph-based approach for HAR, where human joints are treated as graph nodes, and their spatial-temporal relationships are represented as edges, capturing both the spatial structure of the human skeleton and the temporal evolution of the pose over time. However, this model focuses solely on human pose, without accounting for objects that the human may interact with. This limitation makes the ST-GCN less suitable for teleoperating robots in tasks involving instrumental activities of daily living (ADLs), such as food preparation, which require interaction with objects like food and utensils. The interaction between the human and the object can provide valuable spatial relationships for the GCN model to recognize human action. For instance, when a person pushes an object, the object moves in tandem with the hand. Such spatial cues help distinguish actions, like throwing an object, where the object moves away from the person. To better capture these interactions, we incorporate object information into our simplified Spatial-Temporal Graph Convolutional Network (S-ST-GCN).

\begin{figure} [t]
    \centering
    \includegraphics[width=9cm]{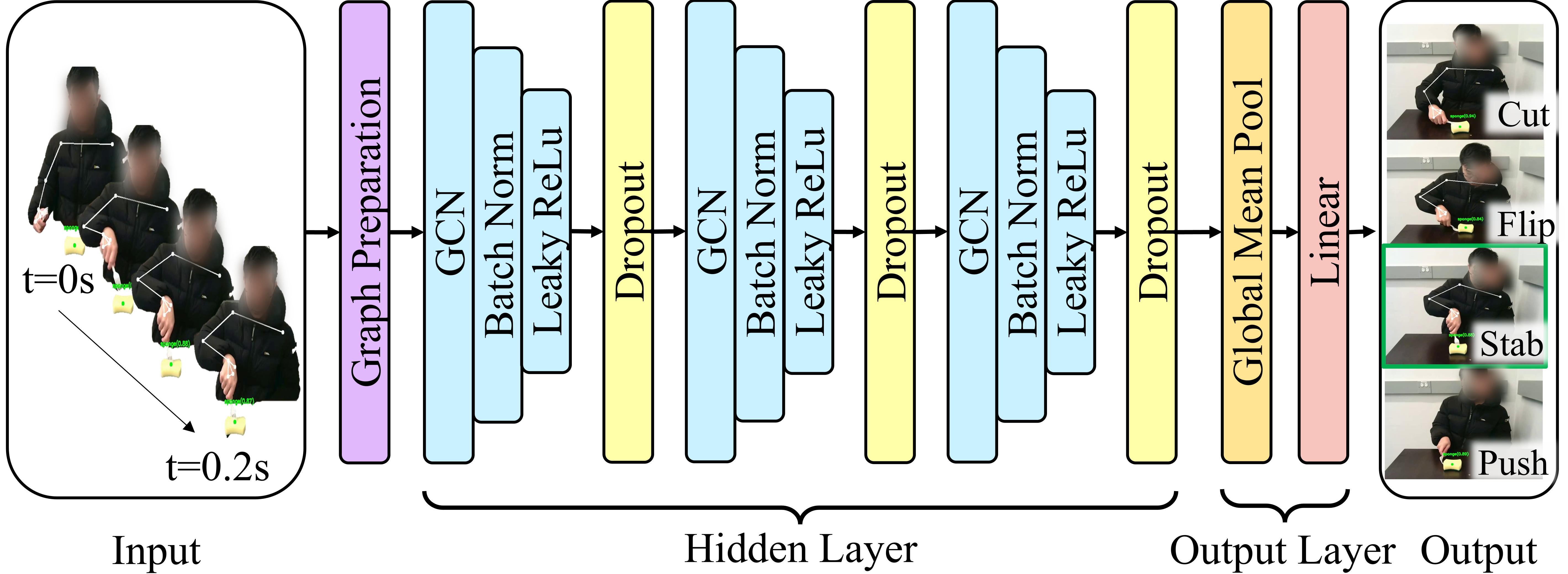}
    \caption{The simplified spatio-temporal Graph Convolutional Network (S-ST-GCN) model for action recognition. The 'Stab' action with a moving window of 0.2s is used as an illustrative example.}
    \label{fig:gnn}
\end{figure}

In this proof-of-concept paper, we evaluated our proposed framework using food preparation tasks, an important instrumental ADL for older adult care. This task was chosen because it is safe and does not require direct interaction with older adult participants during the evaluation of the framework's functionality. As shown in Fig. \ref{fig:gnn}, we selected four key actions related to this task: (1) cut, (2) stab, (3) flip, and (4) push. These actions were chosen as they can be performed with one arm, which is the current focus of our preliminary evaluation. While this framework supports a variety of actions, we acknowledge that more complex actions involving both arms or different movements could be included. Expanding the action space will be possible by employing a more advanced and deeper GCN model, which we plan to explore in future work. For now, we aim to evaluate the feasibility of integrating HAR into teleoperation, and thus, more complex scenarios are beyond the scope of this proof-of-concept study. For the initial experiments, a sponge was used as a proxy for food items, but in future developments, the sponge will be replaced by actual food to better simulate real-world conditions.

Fig. \ref{fig:gnn} illustrates our S-ST-GCN model, a lightweight and effective approach for HAR. Its architecture comprises three GCN layers, each with a hidden dimension of 128, followed by batch normalization and Leaky ReLU activation layers. These three GCN layers capture the local graph structure, with each node receiving information from its 3-hop neighboring nodes. Batch normalization ensures stable feature distributions, while Leaky ReLU activation prevents neurons from becoming inactive. To mitigate overfitting, dropout (rate of 0.3) is applied after each convolutional layer. A global mean pooling layer then aggregates node features into a single graph representation, which is passed through a fully connected linear layer for the classification of the four human actions. This design enables efficient learning of both local and global graph structures while ensuring computational efficiency.

\begin{figure} [t]
    \centering
    \includegraphics[width=5cm]{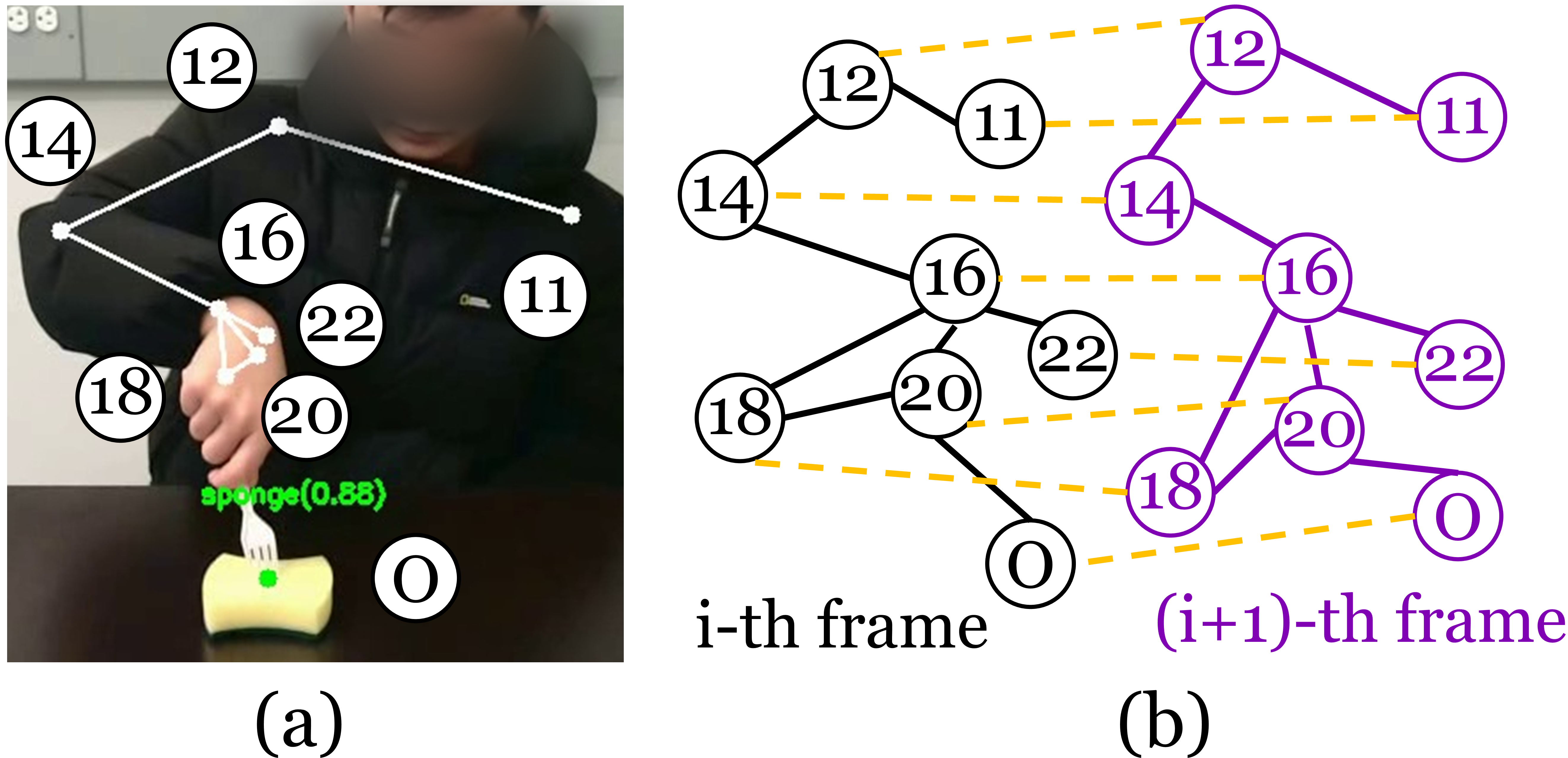}
    \caption{Graph preparation for each frame. (a) Detected pose landmarks and objects. (b) Structures of the spatio-temporal graph. For example, black nodes and edges depict the spatial graph in the i-th frame, while purple nodes and edges represent the graph in the subsequent frame. Yellow lines indicate the temporal edges linking these two frames.}
    \label{fig:gnn_node}
\end{figure}

\begin{figure} [t]
    \centering
    \includegraphics[width=8cm]{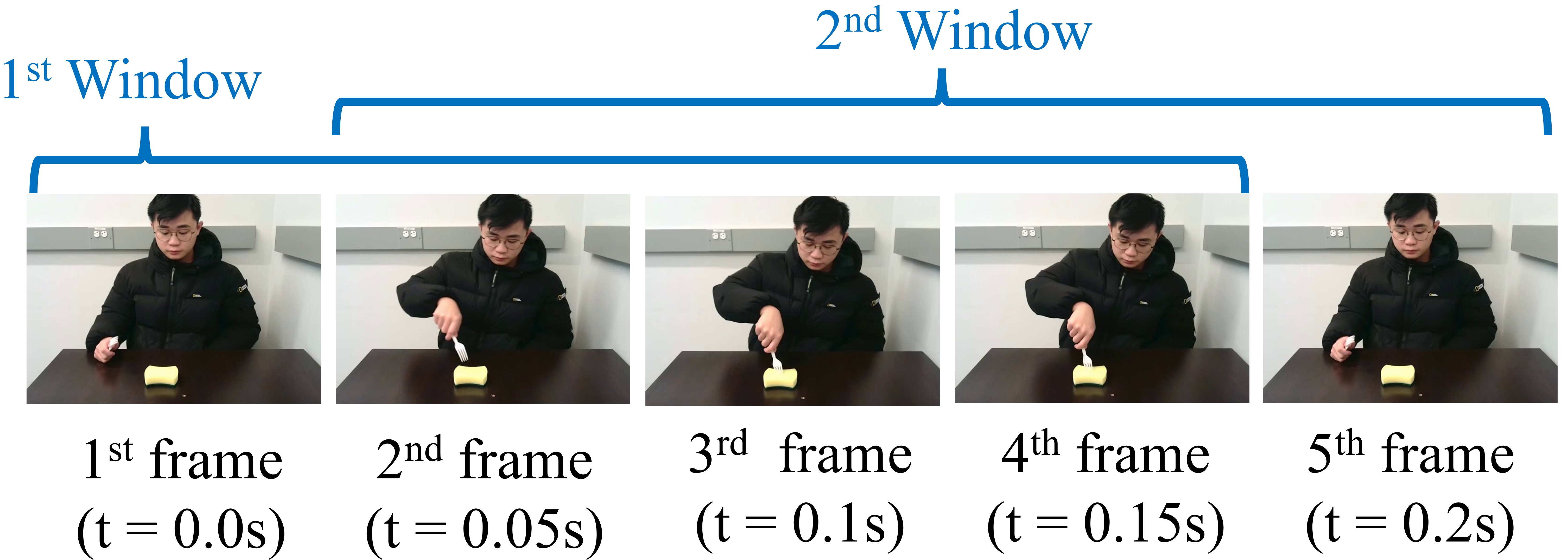}
    \caption{An illustrative example of moving windows with fixed window sizes of 4 frames in a 20 fps video.}
    \label{fig:moving_window}
\end{figure}

As for the graph representation, the node and edge arrangement are depicted in Fig. \ref{fig:gnn_node}. Our S-ST-GCN model contains 8 nodes and 8 spatial edges for each frame. Between the $i$-th and $(i+1)$-th frames, there are 16 nodes ($V_w$) corresponding to pose landmarks ($V_p$) and objects ($V_o$), with 8 temporal edges connecting the corresponding nodes and a total of 16 spatial edges. To enable online action recognition for real-time teleoperation, the action recognition must process continuous data streaming. We employed a moving window ($W_t$) that contains a fixed number ($N_w$) of video frames, as shown in Fig. \ref{fig:moving_window}. Within a moving window, the S-ST-GCN model includes $8N_w$ nodes, $8(N_w-1)$ temporal edges, and $16N_w$ spatial edges. For instance, with an online video feed of 20 fps, a moving window of 4 frames (t = 0.2 s) advances by one frame at each step. This results in the S-ST-GCN in Fig. \ref{fig:gnn} running HAR at 20 Hz. If a more complex GCN is employed, the update frequency may be reduced to accommodate the increased computation time.

For the implementation of S-ST-GCN, we utilized PyTorch Geometric \cite{Fey/Lenssen/2019}. Regarding the nodes, as shown in Fig. \ref{fig:gnn_node}, we employed MediaPipe Pose \cite{48292} to extract pose landmarks for the upper-body skeleton model. Specifically, we focused on key upper-body joint poses based on the Pose Landmark Model (BlazePose \cite{bazarevsky_blazepose_2020}), including the shoulders (11, 12), elbows (14), wrists (16), and fingers (18, 20, 22). For the object node, we fine-tuned a COCO-pretrained YOLOv5 Nano model to detect a sponge, which serves as a proxy for a food item \cite{yolo11_ultralytics}. In the future, we can fine-tune the YOLO model for detecting actual food items.

Once the landmarks and objects are detected, MediaPipe and YOLO compute their 2D coordinates normalized to the frame size (i.e., 640x480 pixels). Subsequently, we translate all landmark and object coordinates relative to the left shoulder (11) to serve as features for each node. This approach ensures that the S-ST-GCN learns relationships based on the landmarks and objects themselves, rather than being influenced by specific pixel locations within the frame.

We followed BlazePose's pose connection to define the spatial edges: (11, 12), (12, 14), (14, 16), (16, 18), (16, 22), (16, 20), and (18, 20). Additionally, we added an edge between the index finger (20) and the object. These nodes and edges effectively capture arm and object movements, as well as their relationships, enabling robust feature extraction for HAR.

Despite these differences from ST-GCN \cite{yan_spatial_2018}, the computation of GCN layers is well-established. For more detailed technical information, the reader can refer to \cite{yan_spatial_2018} and \cite{sanchez_lengeling2021gentle}.

\subsection{Teleoperation Framework with Action Recognition}
Given the challenges in motion-mapping teleoperation mentioned in Section I, we proposed a teleoperation framework using action recognition with S-ST-GCN. In this framework, we aim to remotely control the robot with high-level action commands via recognizing caregivers' motion in real time.

\begin{figure} [t]
    \centering
    \includegraphics[width=8cm]{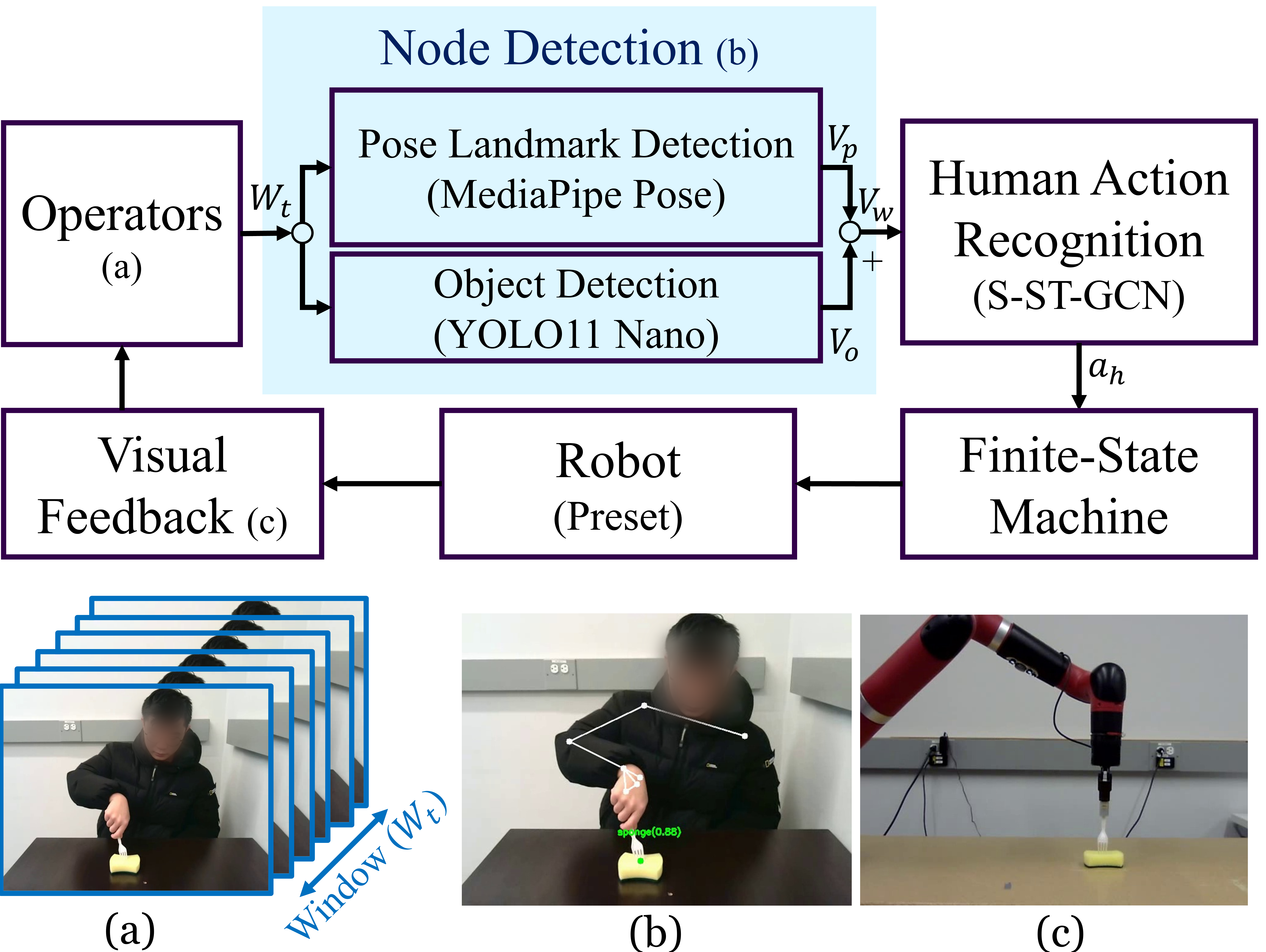}
    \caption{A proposed teleoperation framework using spatio-temporal action recognition. It includes (a) capturing the operator’s movement over a moving window of frames ($W_t$) as an action command, (b) detecting nodes (i.e., pose ($V_p$) and object ($V_o$)) of frames in the moving window ($V_w$) for action recognition, and (c) executing the recognized action ($a_h$) on the robot with visual feedback to the operator.}
    \label{fig:teleopt_framework}
\end{figure}

\begin{figure} [t]
    \centering
    \includegraphics[width=8cm]{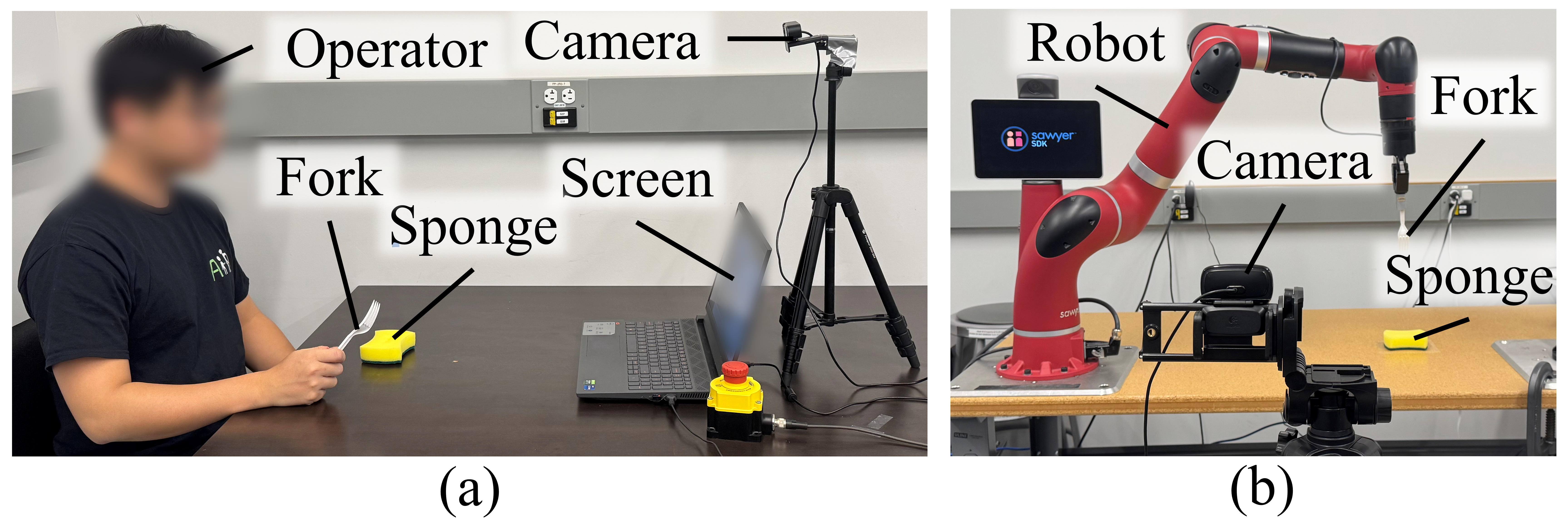}
    \caption{Setup of the teleoperation framework: (a) operator side and (b) remote robot side.}
    \label{fig:gnn_setup}
\end{figure}

Our proposed teleoperation framework is illustrated in Fig. \ref{fig:teleopt_framework}, with the setup shown in Fig. \ref{fig:gnn_setup}. During teleoperation, an RGB camera (Fig. \ref{fig:gnn_setup} (a)) captures a video stream of the operator's actions, as shown in Fig. \ref{fig:moving_window} and Fig. \ref{fig:teleopt_framework} (a). Each frame (Fig. \ref{fig:teleopt_framework} (b)) in the moving window ($W_t$) of the streaming video is used to extract nodes ($V_w$), including nodes for pose landmarks ($V_p$) using MediaPipe Pose, and object nodes ($V_o$) using YOLO11 Nano. The nodes ($V_w$) and spatio-temporal edges are then used to construct the graph for S-ST-GCN. Once the operator's action ($a_h$) is recognized, the robot performs the action. Its motion is captured by another RGB camera (Fig. \ref{fig:gnn_setup} (b)) and sent to the operator's screen as visual feedback, as shown in Fig. \ref{fig:teleopt_framework}(c).

\begin{figure} [t]
    \centering
    \includegraphics[width=6cm]{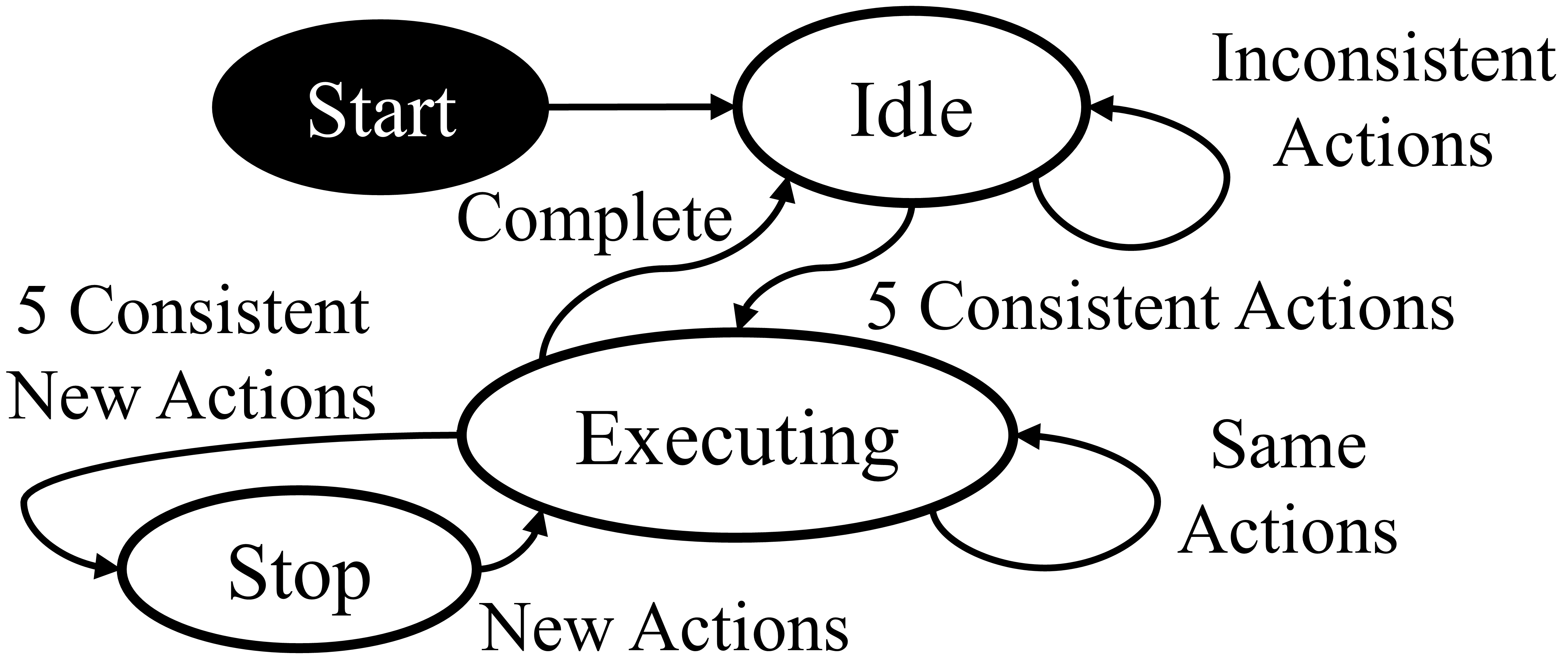}
    \caption{Finite-state machine for robot teleoperation. It illustrates the various robot states corresponding to recognized human actions, particularly those involving inconsistent actions.}
    \label{fig:fsm}
\end{figure}

However, machine learning models are inherently susceptible to errors and may occasionally misclassify actions. For example, a consistent sequence of recognized actions may suddenly shift due to a misclassification. Differentiating between such misclassifications and actual changes in human intent can be challenging, as both result in abrupt transitions in the recognized actions. To address this issue, we incorporate a finite-state machine (FSM) to process recognized actions before execution. As shown in Fig. \ref{fig:fsm}, the FSM helps manage inconsistencies and ensures that the robot responds appropriately, even in the presence of potential recognition errors.

In the FSM, we assume that an actual change in human intent is reflected by a consistent recognition result for at least a short duration, whereas misclassified actions tend to fluctuate abruptly. To address this, the FSM enforces a rule where the robot only executes an action if it is recognized consistently for a predefined number of consecutive frames (e.g., 5 times in a row). Any inconsistent or rapidly changing actions are ignored. Additionally, if the robot is already performing an action, it will only transition to a new one if the new action is consistently recognized; otherwise, it will continue its current task or remain idle. This approach effectively filters out undesired misclassifications, ensuring more stable and reliable robot responses.

As to action execution, as mentioned in Section I, we mainly focus on the feasibility of using action recognition for the teleoperation framework in this proof-of-concept paper. So we use preset robot trajectories to execute the four different actions, although these preset trajectories can be replaced with more advanced motion planning methods, but this is out of the scope of this paper.

\section{Experimental Validation}
\subsection{System Hardware}
The S-ST-GCN training, utilizing PyTorch Geometric, and the fine-tuning of YOLO11 Nano were performed on a Windows laptop equipped with an NVIDIA GeForce GTX 1660 Ti GPU.

For deployment, we implemented the teleoperation framework and human activity recognition (HAR) using S-ST-GCN on a Linux laptop with an 11th Gen Intel Core i7-11800H CPU. Once an action was recognized, the laptop transmitted commands to the Sawyer robot (Rethink Robotics, Boston, U.S.) via Wi-Fi using a real-time ROS publish-subscribe mechanism. Then, robot action was executed through ROS (Intera SDK\footnote{https://github.com/RethinkRobotics/intera\_sdk}), which enabled real-time execution and interaction.

Regarding action recognition, we used the same RGB camera (resolution: 640x480 pixels, frame rate: 20 fps) for both training data collection and deployment.

\subsection{Object Detection with YOLO}
To detect our target object (i.e., sponge) for S-ST-GCN, we used the YOLO11 Nano model (YOLO). Since the COCO dataset does not include sponges, we have fine-tuned the YOLO on our custom sponge dataset with 230 images. 

This custom dataset was prepared using Roboflow \cite{dwyer2024roboflow}. To ensure accuracy, bounding boxes for sponges were annotated using auto-labeling and manual adjustments. The images were randomly divided into 161 for training, 46 for validation, and 23 for testing. The dataset was then exported for offline sponge detection training.

For sponge detection, we fine-tuned the YOLO11 Nano model with 100 training epochs, an initial learning rate of 0.001, a batch size of 16, and the Adaptive Moment Estimation with Decoupled Weight Decay (AdamW) optimizer. After fine-tuning, the model achieved a mAP50-95 of 99.3\% on the validation set and 98.9\% on the test set, demonstrating robust detection and precise object localization for the GNN model.

\subsection{Action recognition with S-ST-GCN}
In this section, we conducted an ablation study to analyze the impact of window size on the HAR with S-ST-GCN. The window size defines the number of frames used in the temporal structure for action recognition. A window that is too small fails to capture sufficient action features, leading to unreliable recognition, while an excessively large window may lead to overfitting and increased computation time, delaying action recognition. To address these trade-offs, we optimize the window size to achieve a balance between recognition accuracy and real-time performance.

To ensure a fair comparison, we trained the S-ST-GCN model with different window sizes using the same process, avoiding any unintended bias. The optimization setup employs the AdamW optimizer with a learning rate of 0.001 and a weight decay of 0.01 to regulate the model and mitigate overfitting. Additionally, a ReduceLROnPlateau learning rate scheduler is implemented, dynamically reducing the learning rate by a factor of 0.5 when validation accuracy stagnates for 10 consecutive epochs, improving convergence stability. The model is trained using cross-entropy loss.

For the dataset, we recorded 280 videos featuring one of the authors, with a resolution of 640×480 pixels and a frame rate of 20 fps. Each video consists of 150 frames (7.5s), resulting in a total of 42.0k frames. To minimize bias, each action class contains an equal number of videos.

Table \ref{tab: acc_window} compares the accuracy across different window sizes. As the window size ($N_w$) increases, the available training sample size decreases, following the relation: $N_s =N_v(N_f-N_w+1)$ for $N_v$ videos with $N_f$ frames each. Initially, larger window sizes improve training, validation, and test accuracy, reaching a peak at 40 frames. Beyond this point, overfitting occurs, where training accuracy remains high while validation and test performance decline. Notably, at 40 frames, the model achieves nearly 90\% accuracy across all action classes in the test set, demonstrating a robust balance between accuracy and generalization. Based on this observation, we adopt a window size of 40 frames for teleoperation evaluation in Section III. D.

\begin{table}[htbp]
\centering
\caption{Accuracy Comparison of Different Window Sizes}
\vspace{-0.5em}
\resizebox{8.5cm}{!}{ 
\begin{tabular}{|c|c|c|c|c|c|c|c|c|}
\hline
\textbf{${N_w}$$^{\mathrm{a}}$} & \textbf{${N_s}$$^{\mathrm{b}}$} & \textbf{Train} & \textbf{Valid} & \multicolumn{5}{c|}{\textbf{Test (\%)}} \\
\cline{5-9}
\textbf{} & \textbf{} & \textbf{(\%)} & \textbf{(\%)} & \textbf{All} & \textbf{Cut} & \textbf{Flip} & \textbf{Stab} & \textbf{Push} \\

\hline
1& 42.0k& 76.7& 78.5& 83.6& 84.7& 83.1& 86.2& 80.3 \\
2& 41.7k& 77.2& 80.7& 83.1& 83.0& 83.3& 86.7& 79.3 \\
5& 40.9k& 84.7& 84.1& 84.8& 86.9& 78.6& 88.8& 84.9 \\
10& 39.5k& 86.0& 85.6& 84.5& 87.4& 75.2& 91.3& 84.0 \\
20& 36.7k& 91.5& 91.1& 89.3& 87.8& 87.6& 96.3& 85.7 \\
40& 31.1k& 97.0& 94.7& 92.3& 92.5& 89.2& 100.00& 87.4 \\
60& 25.5k& 98.4& 93.7& 92.1& 91.5& 91.2& 100.00& 85.7 \\
80& 19.9k& 97.5& 92.0& 92.4& 90.7& 93.0& 100.00& 85.7 \\
150& 280& 82.1& 80.7& 82.8& 85.7& 57.1& 100.00& 87.5 \\
\hline
\multicolumn{9}{l}{$^{\mathrm{a}}$${N_w}$: No. of frames in a window.} \\
\multicolumn{9}{l}{$^{\mathrm{b}}$${N_s}$: Sample size.}
\end{tabular}
}
\label{tab: acc_window}
\end{table}

\begin{figure} [t]
    \centering
    \includegraphics[width=8.5cm]{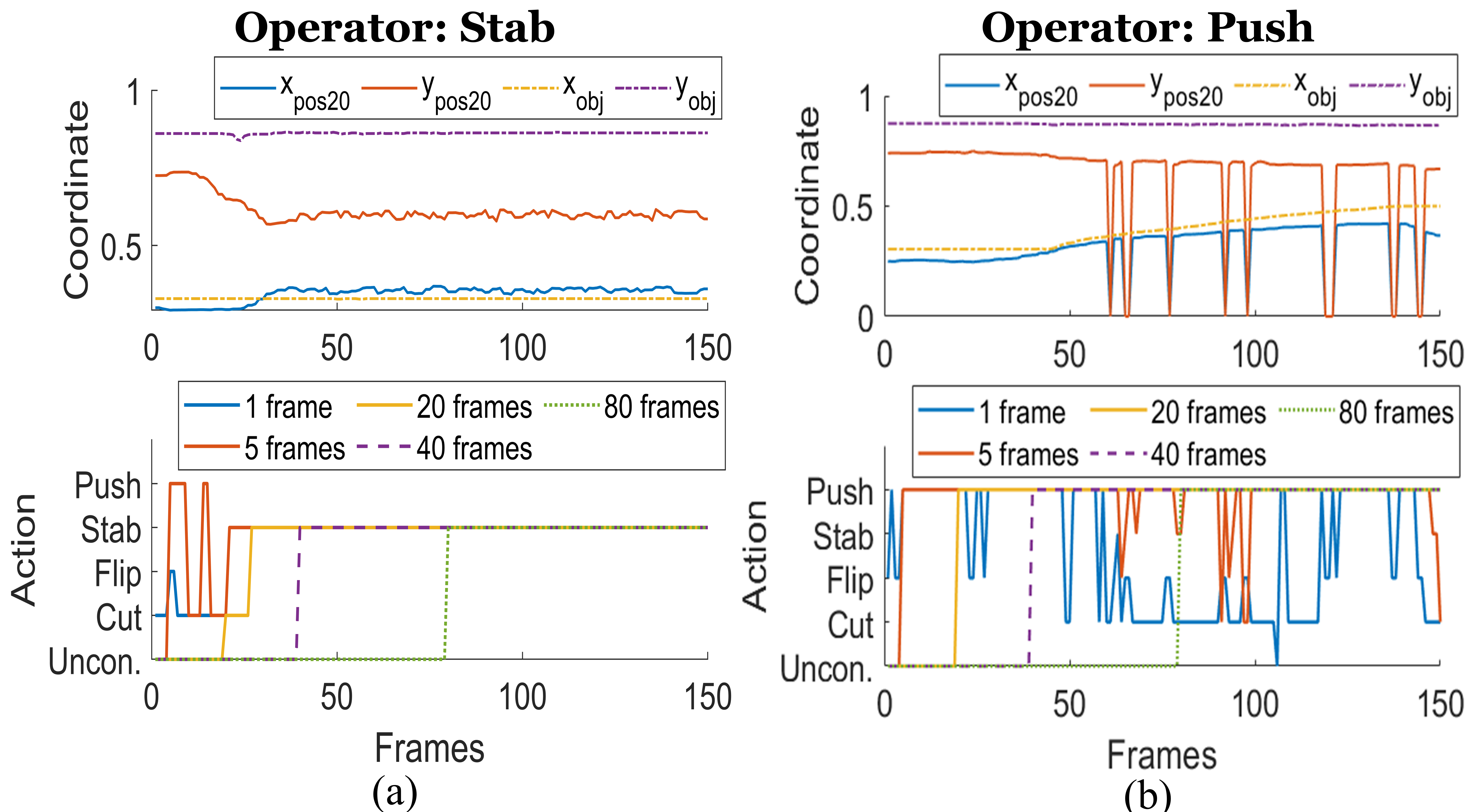}
    \caption{The effect of window sizes on action recognition performance. An operator performed (a) "stab" and (b) "push" actions. The upper section presents the normalized coordinates in pixel space for each frame. $x_{pos20}$ and $y_{pos20}$ denote right index finger coordinates, while $x_{obj}$ and $y_{obj}$ indicate the sponge coordinates. The lower section shows the recognized actions with different window sizes in frames.}
    \label{fig:classify_result}
\end{figure}

Fig. \ref{fig:classify_result} illustrates the effect of window size on action recognition accuracy and classification consistency. Smaller window sizes (e.g., 1 and 5 frames) result in high variability, particularly during the first 20 frames in the "stab" action. This inconsistency is likely due to the limited temporal context available for recognition. Additionally, poor video quality may cause pose landmarks ($x_{pos20}$ and $y_{pos20}$) to disappear, as seen in Figure \ref{fig:classify_result} (b) for "push" action, further degrading recognition performance for shorter windows. 

Although larger window sizes improve recognition accuracy, they introduce an initial delay in robot action execution. However, this delay does not affect teleoperation performance after initialization. For instance, with an 80-frame window and a 20 fps video stream, the system requires a 4-second waiting period before recognizing the first action. Nevertheless, since the system employs a moving window approach (shifting forward by one frame at a time, as illustrated in Fig. \ref{fig:moving_window}), the recognition updates at 20 Hz, ensuring continuous and real-time teleoperation without further delays.

\begin{figure} [t]
    \centering
    \includegraphics[width=8cm]{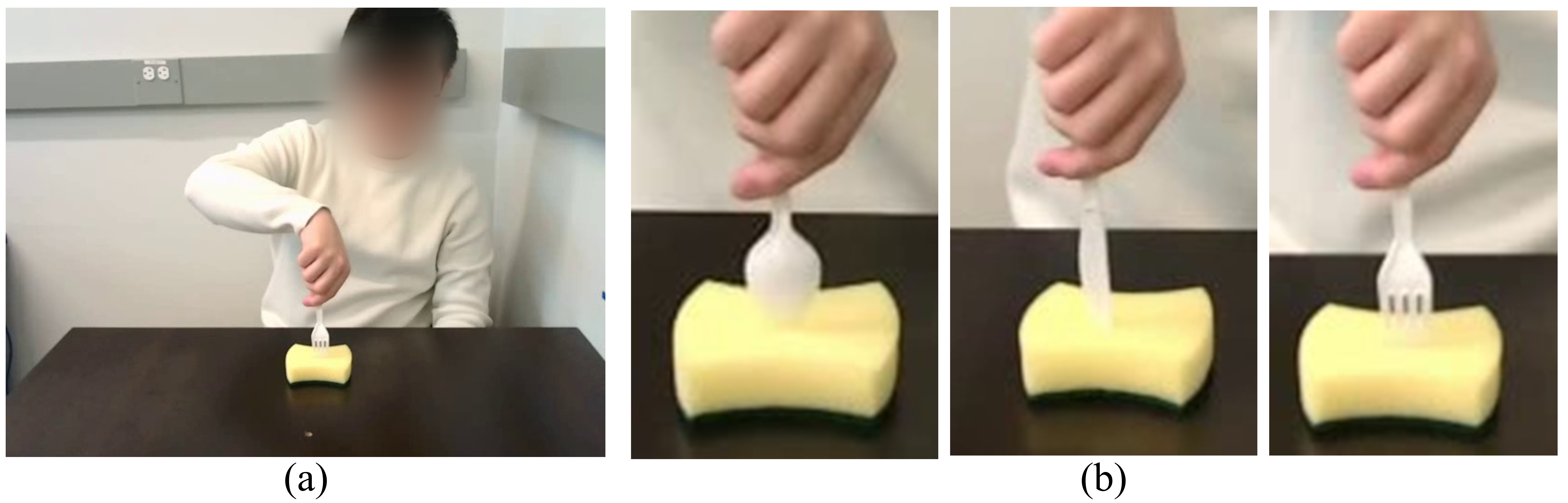}
    \caption{(a) Camera view after applying environmental changes, including slight camera adjustments and variations in the operator's attire. (b) Three different utensil types: spoon (L), knife (M), and fork (R), with the spoon and knife being unseen utensils.}
    \label{fig:utentil_classify}
\end{figure}

To evaluate the robustness of the GNN model, we introduced various environmental changes and tested its accuracy. Specifically, we tested the model with different utensils (Fig. \ref{fig:utentil_classify}): a fork, a spoon, and a knife, and recorded the model accuracy shown in Table \ref{tab: acc_utensil}. Notably, the spoon and knife were considered unseen objects during training. In addition to utensil variation, we introduced further environmental changes by moving the camera backward by approximately 5 cm and altering the operator's clothing, as shown in Fig. \ref{fig:utentil_classify} (a). Despite these modifications, the GNN model demonstrated exceptional reliability and robustness, achieving an overall accuracy exceeding 92.0\% and maintaining at least 81.3\% accuracy across all individual actions with different utensils. These results underscore the model’s ability to focus primarily on node features, i.e., the coordinates of pose landmarks and sponge, rather than environmental factors like utensil type, clothing, or camera position. This robustness is particularly advantageous for real-world applications, where factors such as operator attire, utensil variation, or slight camera adjustments can change without necessitating model retraining.

\begin{table}[htbp]
\centering
\caption{Accuracy Comparison for Different Utensil Types}
\vspace{-0.5em} 
\begin{tabular}{|c|c|c|c|c|c|}
\hline
\textbf{Utensil Type} & \textbf{All} & \textbf{Cut} & \textbf{Flip} & \textbf{Stab} & \textbf{Push} \\

\hline
Fork& 97.3& 97.2& 100.0& 96.0& 95.9 \\
Spoon (Unseen) & 96.6& 96.7& 94.5& 100.0& 95.1 \\
Knife (Unseen) & 92.0& 88.4& 98.8& 99.5& 81.3 \\

\hline
\end{tabular}
\label{tab: acc_utensil}
\end{table}

\subsection{Teleoperation Performance}
This section evaluates teleoperation performance using action recognition and assesses the functionality of the finite state machine (FSM) in handling misclassified actions. The FSM ensures that the robot executes an action only if it is consistently recognized over a predefined number of consecutive frames. In the following experiment, we implemented the S-ST-GCN model with a window size of 40 frames to analyze its effectiveness in action recognition and teleoperation stability.

\begin{figure} [t]
    \centering
    \includegraphics[width=8cm]{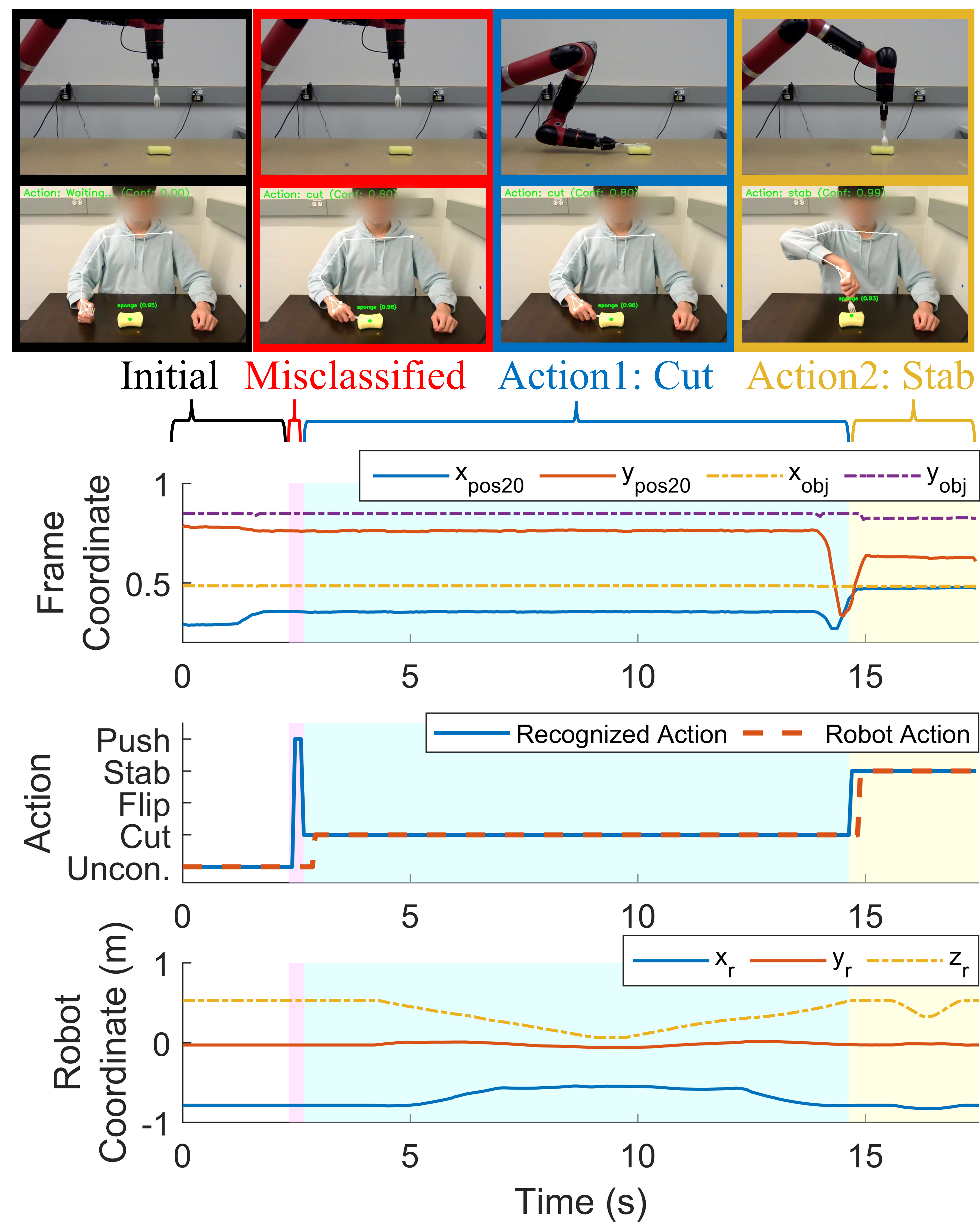}
    \caption{Teleoperation performance with S-ST-GCN action recognition. The upper section of the plot displays normalized coordinates, where $x_{pos20}$ and $y_{pos20}$ represent the right index finger coordinates, and $x_{obj}$ and $y_{obj}$ denote the sponge coordinates. The middle section illustrates recognized actions and executed robot actions. The lower section presents the robot's endpoint coordinates: $x_r$, $y_r$, and $z_r$.}
    \label{fig:teleopt_result}
\end{figure}

As shown in Fig. \ref{fig:teleopt_result}, the robot required a 2-second waiting period before recognizing the first action. Following this, a misclassified action occurred; however, the FSM effectively handled it, preventing unintended execution. Once the first valid action, "Cut," was consistently recognized, the robot executed it approximately 2.0 seconds after the operator completed the action. This delay is expected due to the initial waiting period, the duration of misclassification, and potential Wi-Fi communication latency between the laptop and the robot via the router. When the operator transitioned to the "Stab" action, the FSM correctly identified this as an intentional action rather than a misclassification and executed "Stab" with a short delay of 0.4 seconds after the operator completed the motion. 


Furthermore, Fig. \ref{fig:teleopt_result} also highlights differences in the speed and posture of human and robot movements. This indicates that once the robot receives an action command, it autonomously executes the motion without direct human control. Consequently, the operator can perform natural movements without worrying about exceeding velocity limits or colliding with the environment.

In this experiment, we invited another author to perform the teleoperation without retraining the S-ST-GCN model. The results demonstrate the model's robustness, as it accurately recognized actions regardless of the operator or attire, further confirming its adaptability to different users and environmental conditions.

\section{Discussion}

This study demonstrated that S-ST-GCN action recognition offers a promising alternative to traditional motion-mapping teleoperation, which often requires operators to slow down, adjust postures, or remain stationary to synchronize with the robot due to hardware constraints and safety concerns. These limitations lead to unnatural movements and muscle fatigue \cite{lin_physical_2019}, particularly in complex tasks like ADLs, increasing caregiver burden and posing occupational health risks. 
In contrast, our framework allows operators to move effortlessly without being constrained by the robot’s execution speed. By dynamically recognizing human actions, the system eliminates the need for human-robot synchronization during teleoperation, potentially reducing operator fatigue and improving usability in long-term old adult care.

A key advantage of our approach is its ease of use and minimal setup requirements. Unlike motion-mapping teleoperation, which necessitates precise coordinate calibration between the human and robot, our system eliminates this need, without the need for kinematic computations or precise human motion tracking. This makes it easier to generalize implementation across different robotic platforms. Additionally, operators do not need to wear sensors or markers, significantly reducing setup time and making the system more accessible and convenient for caregivers. While recognition accuracy depends on consistent camera placement, this limitation can be addressed by expanding the dataset to include various camera orientations and positions. Future work will focus on enhancing model generalization by incorporating a broader range of actions, diverse human demonstrations, and different participants.

Beyond the proposed teleoperation framework with preset robot trajectory, this framework has great potential to integrate with more intelligent and adaptive robotic behaviors. By leveraging advanced motion planning techniques such as reinforcement learning with diffusion policy learning from demonstration \cite{wang_diffusion_2022, wang_diffusion_2023, chi_diffusion_2024}, robots could assist users more effectively. In this case, operators are not required to precisely control remote robot motion to complete ADLs, making the system more user-friendly. This is particularly valuable in scenarios where operators have a limited field of view of the robot camera and may not fully perceive environmental hazards. A semi-autonomous robot under a shared control framework would enhance safety by preventing unintended collisions with the older adults or the surrounding environment.

At the current stage of development, our method successfully enables teleoperation with action recognition. However, a limitation is that the first recognized action is occasionally misclassified, likely because the operator is unaware of when action recognition begins. Future work will address this issue by enhancing the graphical user interface to provide clear visual cues indicating the start of action recognition.

Additionally, this proof-of-concept study focuses on developing and evaluating the proposed teleoperation framework. While our results demonstrate its potential, we cannot yet conclude that this approach achieves telepresence or outperforms traditional motion-mapping teleoperation in terms of key metrics such as user workload and perceived usability. To further validate its effectiveness, we plan to conduct user studies in future research.

\section{Conclusion}

This paper demonstrates the feasibility of integrating action recognition into a robot teleoperation framework, specifically designed to assist caregivers in performing activities of daily living (ADLs) for older adults. Using our simplified Spatio-Temporal Graph Convolutional Network (S-ST-GCN) model, we successfully recognize human actions and enable the robot to execute corresponding preset trajectories. This framework offers a more natural and intuitive control mechanism, eliminating the need for motion synchronization between human and robot, and potentially reducing operator fatigue. Additionally, we introduce a finite-state machine (FSM) to address misclassifications of human actions, ensuring reliable and accurate robot behavior. As a proof-of-concept, this study highlights the promise of our framework in performing ADLs for older adults, potentially supporting their independent living. Future work will focus on improving the model's generalization by incorporating additional actions, exploring advanced motion planning techniques to enhance the robot’s flexibility in ADL assistance, and conducting user studies to assess the telepresence and ease of control.

\bibliographystyle{IEEEtran}
\bibliography{references}

\vspace{12pt}

\end{document}